\def\isarxiv{1} 

\ifdefined\isarxiv
\documentclass[11pt]{article}

\usepackage[numbers]{natbib}

\else
\documentclass{article}
\usepackage{microtype}
\usepackage{graphicx}
\usepackage{subfig}
\usepackage{hyperref}
\usepackage{icml2024}
\fi

\usepackage{amsmath}
\usepackage{amsthm}
\usepackage{amssymb}
\usepackage{algorithm}
\usepackage{algpseudocode}
\usepackage{grffile}
\usepackage{wrapfig,epsfig}
\usepackage{url}
\usepackage{xcolor}
\usepackage{epstopdf}

\usepackage{bbm}
\usepackage{dsfont}

\allowdisplaybreaks

\ifdefined\isarxiv

\usepackage{tikz}
\usepackage{hyperref}  
\hypersetup{colorlinks=true,citecolor=blue,linkcolor=blue} 
\usetikzlibrary{arrows}
\usepackage[margin=1in]{geometry}

\else

\fi

\newtheorem{theorem}{Theorem}[section]
\newtheorem{lemma}[theorem]{Lemma}
\newtheorem{definition}[theorem]{Definition}

\newtheorem{fact}[theorem]{Fact}
\newtheorem{remark}[theorem]{Remark}

\newtheorem{hypothesis}[theorem]{Hypothesis}

\newcommand{\wt}{\widetilde}

\newcommand{\N}{\mathcal{N}}
\newcommand{\R}{\mathbb{R}}

\renewcommand{\d}{\mathrm{d}}

\renewcommand{\tilde}{\wt}

\newcommand{\Tmat}{{\cal T}_{\mathrm{mat}}}
\newcommand{\diag}{\mathrm{diag}}

\DeclareMathOperator{\poly}{poly}
\DeclareMathOperator{\A}{\mathsf{A}}
\DeclareMathOperator{\SETH}{\mathsf{SETH}}

\DeclareMathOperator{\vect}{vec}

\makeatletter
\newcommand*{\RN}[1]{\expandafter\@slowromancap\romannumeral #1@}
\makeatother
\newcommand{\Zhao}[1]{{\color{red}[Zhao: #1]}} 
\newcommand{\Josh}[1]{{\color{blue}[Josh: #1]}} 

\renewcommand{\Zhao}[1]{{}} 
\renewcommand{\Josh}[1]{{}} 

\usepackage{lineno}

\ifdefined\isarxiv

\else
\usepackage[textsize=tiny]{todonotes}

\icmltitlerunning{The Fine-Grained Complexity of Gradient Computation for Training Large Language Models}
\fi

\begin{document}

\ifdefined\isarxiv

\date{}

\title{The Fine-Grained Complexity of Gradient Computation for Training Large Language Models}
\author{
Josh Alman\thanks{\texttt{josh@cs.columbia.edu}. Columbia University.}
\and
Zhao Song\thanks{\texttt{zsong@adobe.com}. Adobe Research.}
}

\else

\twocolumn[
\icmltitle{The Fine-Grained Complexity of Gradient Computation for Training Large Language Models}



\icmlsetsymbol{equal}{*}

\begin{icmlauthorlist}
\icmlauthor{Firstname1 Lastname1}{equal,yyy}
\icmlauthor{Firstname2 Lastname2}{equal,yyy,comp}
\icmlauthor{Firstname3 Lastname3}{comp}
\icmlauthor{Firstname4 Lastname4}{sch}
\icmlauthor{Firstname5 Lastname5}{yyy}
\icmlauthor{Firstname6 Lastname6}{sch,yyy,comp}
\icmlauthor{Firstname7 Lastname7}{comp}
\icmlauthor{Firstname8 Lastname8}{sch}
\icmlauthor{Firstname8 Lastname8}{yyy,comp}
\end{icmlauthorlist}

\icmlaffiliation{yyy}{Department of XXX, University of YYY, Location, Country}
\icmlaffiliation{comp}{Company Name, Location, Country}
\icmlaffiliation{sch}{School of ZZZ, Institute of WWW, Location, Country}

\icmlcorrespondingauthor{Firstname1 Lastname1}{first1.last1@xxx.edu}
\icmlcorrespondingauthor{Firstname2 Lastname2}{first2.last2@www.uk}

\icmlkeywords{Machine Learning, ICML}

\vskip 0.3in
]



\printAffiliationsAndNotice{\icmlEqualContribution} 

\fi

\ifdefined\isarxiv
\begin{titlepage}
  \maketitle
  \begin{abstract}
Large language models (LLMs) have made fundamental contributions over the last a few years. To train an LLM, one needs to alternatingly run `forward' computations and `backward' computations. The forward computation can be viewed as attention function evaluation, and the backward computation can be viewed as a gradient computation. In previous work by [Alman and Song, NeurIPS 2023], it was proved that the forward step can be performed in almost-linear time in certain parameter regimes, but that there is no truly sub-quadratic time algorithm in the remaining parameter regimes unless the popular hypothesis $\SETH$ is false. In this work, we show nearly identical results for the harder-seeming problem of computing the gradient of loss function of one layer attention network, and thus for the entire process of LLM training. This completely characterizes the fine-grained complexity of every step of LLM training.

  \end{abstract}
  \thispagestyle{empty}
\end{titlepage}

{\hypersetup{linkcolor=black}
\tableofcontents
}
\newpage

\else

\begin{abstract}
Large language models (LLMs) have made fundamental contributions over the last a few years. To train an LLM, one needs to alternatingly run `forward' computations and `backward' computations. The forward computation can be viewed as attention function evaluation, and the backward computation can be viewed as a gradient computation. In previous work by [Alman and Song, NeurIPS 2023], it was proved that the forward step can be performed in almost-linear time in certain parameter regimes, but that there is no truly sub-quadratic time algorithm in the remaining parameter regimes unless the popular hypothesis $\SETH$ is false. In this work, we show nearly identical results for the harder-seeming problem of computing the gradient of loss function of one layer attention network, and thus for the entire process of LLM training. This completely characterizes the fine-grained complexity of every step of LLM training.
\end{abstract}

\fi

\section{Introduction}

Large language models (LLMs) have emerged as popular technologies, driving breakthroughs across many applications in natural language processing, computer vision, translation, and many other areas     \cite{vsp+17,dclt18,log+19,ydy+19,bmr+20,zrg+22,cnd+22,tli+23,tms+23,adobe_firefly,m23,tdh+22,ycri22,wtb+22}. The training of these models is a computationally intensive process, characterized by alternating between two primary operations: forward computation and backward computation. Forward computation, or function evaluation, involves the propagation of input data through the network to generate predictions. Conversely, backward computation, or gradient computation, is the process of calculating the gradient of the loss function with respect to the model's parameters, facilitating the optimization of these parameters during training.

The efficiency of these computations directly impacts the feasibility and scalability of training LLMs, particularly as models grow in size and complexity. Recent work by \cite{as23,as24} has carefully studied the \emph{forward} computation step. They demonstrated a sharp computational boundary, showing that how quickly the forward steps can be performed depends critically on how large the entries are of the matrices which define the model parameters. They showed a near-linear time algorithm when these entries are small, and also proved that when the entries are large, there is no algorithm much faster than the trivial algorithm, contingent upon the Strong Exponential Time Hypothesis ($\SETH$) \cite{ip01} holding true. This finding underscores a fundamental limitation in accelerating the training of LLMs, raising pivotal questions about the inherent computational complexity of these models.

The Strong Exponential Time Hypothesis ($\SETH$) was introduced by Impagliazzo and Paturi \cite{ip01} over 20 years ago. It is a strengthening of the $\mathsf{P} \neq \mathsf{NP}$ conjecture, and asserts that our current best $\mathsf{SAT}$ algorithms are roughly optimal (for detailed statement, see Hypothesis~\ref{hyp:seth} below). $\SETH$ is a popular conjecture from fine-grained complexity theory which has been used to prove lower bounds for a wide variety of algorithmic problems. See, for instance, the survey~\cite{w18}.

In other words, in some parameter regimes, the algorithm of~\cite{as23} performs the forward steps about as quickly as one could hope for, whereas in other regimes, assuming $\SETH$, it is impossible to design a nontrivially fast algorithm. However, this leaves open many important questions about LLM training. In the case when forward computation can be done quickly, can the same be said for backward computation? If not, then the entire training process would still be slow. Relatedly, in parameter regimes where forward computation is known to be hard, is backward computation also hard? If not, perhaps heuristic tricks could be used, or other details of the model could be modified, to speed up the overall training. As we will see shortly, the backward step is defined in a much more complicated way than the forward step, and it is not evident that algorithms or lower bounds for one extend to the other.

Our study aims to resolve these questions and determine the fine-grained complexity of the backward computation phase. Our main result (which we state more foramlly shortly) shows that the same computational threshold from forward computation also arises for the backward problem, and that the problems are easy (opr hard) in the exact same parameter regimes. Thus, the forward algorithm of~\cite{as23} can be combined with our novel backward algorithm to perform each training step for LLMs in near-linear time when the parameter matrix entries are small enough, whereas when the entries are not small enough, neither step can be performed quickly.

In addition to characterizing the fine-grained complexity of LLM training, our result for gradient computation is novel for a few reasons.
\begin{itemize}
    \item Previous work on computational lower bounds, only focuses on forward computation, see \cite{as23,kwh23,as24}. To our knowledge, ours is the first work to prove hardness of a backward computation step for training an LLM or similar model.
    \item There has been previous work on the algorithms for backward/gradient computation \cite{bpsw21,syz21,hswz22,als+23,gqsw24,szz24}. That said, most of these works focus on backwards computation in other settings. The only previous work we're aware of that studies the optimization of attention layers (for LLMs) is \cite{gswy23}, which uses Newton method that rely on Hessian computation. However, Hessian computation  is substantially more expensive than gradient computation; our results apply to the gradient computation and get around the Hessian ``barrier'', allowing for faster algorithms in some parameter regimes, and more powerful lower bounds in others.
\end{itemize}

\subsection{Problem Definition}

Before formally stating our results, we begin by precisely defining the problems we study. We begin with the following problem of the computation 
of general Attention forward layer.

\begin{definition}[$\ell$-th layer forward computation]
Given weights $Q,K, V \in \R^{d \times d}$, and letting $E_{\ell} \in \R^{n \times d}$ denote the $\ell$-th layer input, then $E_{\ell+1} \in \R^{n \times d}$ is defined recursively as
\begin{align*}
 E_{\ell+1} \gets D^{-1} \exp( E_{\ell} Q K^\top E_{\ell}^\top / d) E_{\ell} V
\end{align*} 
where 
\begin{itemize}
\item
$D:= \diag( \exp( E_{\ell} Q K^\top E_{\ell}^\top / d){\bf 1}_n )$. 
\item $\exp$ denotes the exponential function which is entry-wise, i.e., $\exp(A)_{i, j} = \exp(A_{i, j})$ for all matrices $A$.
\item $\diag()$ operation takes a vector as input and generates a diagonal matrix with the entries of that vector.
\item ${\bf 1}_n$ denotes the length-$n$ all ones vector.
\end{itemize}
\end{definition}

In mathematical terms, optimization in the context of attention computation is described as (by renaming the $QK^\top \in \R^{d \times d}$ to be $X \in \R^{d \times d}$ and $V \in \R^{d \times d}$ to be $Y \in \R^{d \times d}$):
\begin{definition}[Attention optimization]\label{def:attention_optimization_loss}
    Given four $n \times d$ size matrices $A_1, A_2, A_3$ and $E \in \R^{n \times d}$. Suppose that a $d \times d$ size square matrix $ Y \in \R$ is also given. The attention optimization problem is formulated as:
    \begin{align*}
        \min_{X \in \R^{d \times d}}  L(X) := 0.5 \| D(X)^{-1} \exp(A_1 X A_2^\top/d) A_3 Y - E \|_F^2.
    \end{align*}
    Here $D(X) \in \R^{n \times n}$ is 
    \begin{align*}
        D(X) := \diag( \exp(A_1 X A_2^\top /d ) {\bf 1}_n ).
    \end{align*}
    and $\| \cdot \|_F^2$ denotes the squared Frobenius norm, i.e., $\| A \|_F^2:= \sum_{i,j} A_{i,j}^2$.
\end{definition}

\begin{remark}
In principle, the loss function above, and resulting gradients below, should depend on both $X$ and $Y$. However, since the final matrix computed in the norm in $L$ depends only linearly on $Y$, it is straightforward to incorporate it into either an algorithm or lower bound.
Thus, in this work, we focus on the case where $X$ is variable and $Y$ is a fixed input to simplify some arguments.
\end{remark}

We thus define Approximate Attention Loss function Gradient Computation problem as follows:
\begin{definition}[Approximate Attention Loss Gradient Computation ($\mathsf{AAttLGC}(n,d,\epsilon)$)]\label{def:AAttLGC}
Given four $n \times d$ size matrices $A_1 \in \R^{n \times d}, A_2 \in \R^{n \times d}, A_3 \in \R^{n \times d},$, $E \in \R^{n \times d}$ and a square matrix $ Y \in \R^{d \times d}$ to be fixed matrices. Assume that $\| A_1 X \|_{\infty} \leq B$, $ \|A_2 \|_{\infty} \leq B$. Assume all numbers (in matrices) are also in $\log(n)$ bits model. 
Let $L(X)$ be defined as Definition~\ref{def:attention_optimization_loss}.
Let $\frac{\d L(X)}{\d X}$ denote the gradient of loss function $L(x)$. 

The goal is to output a vector $\wt{g}$ such that
\begin{align*}
    \| \wt{g} - \frac{ \d L(X) }{ \d X } \|_{\infty} \leq \epsilon.
\end{align*}
Here for matrix $A$, $\| A \|_{\infty}:= \max_{i,j} |A_{i,j}|$.
\end{definition}

\subsection{Main Results}

Our main results show that there is a threshold in the computational complexity of $\mathsf{AAttLGC}(n, d = O(\log n))$ depending on the bound $B$. When $B= o(\sqrt{\log n} )$ we give a new near-linear-time algorithm, and when $B= \omega(\sqrt{\log n} )$, we show that such an algorithm is impossible assuming SETH. This matches the results of~\cite{as23}, where a nearly identical threshold at $B$ around $\sqrt{\log n}$ was also observed. Our results therefore imply that the entire LLM training process has this computational threshold.

\begin{theorem}[Main result, Lower bound, informal version of Theorem~\ref{thm:mainlb:formal}] \label{thm:mainlb}
Assuming $\mathsf{SETH}$, there is no algorithm running in time $O(n^{2-q})$ for any $q>0$ for the $\mathsf{AAttLGC}(n, d = O( \log n ), B = \omega(\sqrt{\log n} ))$ (see Definition~\ref{def:AAttLGC}).
\end{theorem}

\begin{theorem}[Main result, Upper bound, informal version of Theorem~\ref{thm:mainalg:formal}] \label{thm:mainalg}
Assuming entries are bounded, there is a $n^{1+o(1)}$ time algorithm to solve $\mathsf{AAttLGC}(n, d = O( \log n), B = o(\sqrt{\log n} ))$ (see Definition~\ref{def:AAttLGC}) up to $1/\poly(n)$ accuracy.
\end{theorem}

Our new algorithm (Theorem~\ref{thm:mainalg}) builds on a low-rank approximation for the attention matrix from prior work~\cite{aa22,as23}. Incorporating these approximation into the gradient computation is not straightforward; in the forward problem, one simply multiplies the attention matrix by an input value matrix, but in the backward problem, it is combined with other matrices in an intricate (non-linear) way. We ultimately use tools from tensor algebra to get a handle on the entry-wise products and high-rank sparse matrices which arise in the gradient computation but do not typically preserve the needed low-rank structure.

Our new lower bound (Theorem~\ref{thm:mainlb}) comes from a careful reduction from a special case the forward problem (where hardness is known from prior work) to the backward problem. Reducing from computing a function to computing its gradient in general is quite challenging or impossible without control over how quickly the gradient may be growing or changing, and in general, the gradient of the forward (attention) computation can behave quite erratically (which is likely necessary for the expressive power of attention units). Nonetheless, in the special case of the inputs for which attention computation is known to be hard from prior work, we are able to reasonably control the growth of these gradients and successfully perform our reduction.

{\bf Roadmap.}
We discuss other related works in Section~\ref{sec:related}. In Section~\ref{sec:preli}, we provide the basic notation, definitions, backgrounds, and facts which we will use. In Section~\ref{sec:upper_bound}, we provide the proof sketch of our algorithm and defer the details to the Appendix. In Section~\ref{sec:lower_bound}, we provide our main lower bound result. In Section~\ref{sec:conclusion}, we briefly conclude our paper.  

\section{Related Work}\label{sec:related}

\paragraph{Fine-grained Complexity}

Numerous algorithmic techniques have been used in theory and in practice for attention computations. The first algorithm with provable guarantees, by Zandieh, Han, Daliri, and Karbasi~\cite{zhdk23}, used locality sensitive hashing (LSH) techniques \cite{ckns20}, while later work by Alman and Song~\cite{as23} used polynomial approxmation methods~\cite{acss20,aa22}. We particularly focus here on the latter technique, which is the only algorithm we're aware of which achieves near-linear running time.

Keles, Wijewardena, and Hedge~\cite{kwh23} established the first lower bound on attention computation under the assumption of $\mathsf{SETH}$. Their findings demonstrated that when $d = \omega(\log n)$, it is not possible to execute forward computations in subquadratic time. The later lower bound of \cite{as23} further incorporated the magnitudes of the input entries into the lower bound to tightly match the aforementioned algorithms. Both use the high-level technique of~\cite{bis17} from kernel density estimation, and build on methods derived from fine-grained complexity associated with approximate nearest neighbor search~\cite{r18} and the polynomial method~\cite{aa22}.

\paragraph{Fast Attention Computation}

Optimizing the computation of attention mechanisms in pre-trained LLMs, given their extensive parameter sets, has been a focal point of recent research. Various studies have explored the application of locality sensitive hashing (LSH) techniques to approximate attention mechanisms. \cite{kkl20} introduced two methods to enhance computational efficiency, including the use of LSH to replace dot product attention and a reversible residual layer to substitute the standard residual layer. \cite{clp+21} refined this approximation, noting that LSH's efficiency does not require constant parameter updates. \cite{zhdk23} proposed an innovative estimator based on Kernel Density Estimation (KDE) to speed up the softmax function and matrix multiplication computations. Some recent works~\cite{hjk+23,kmz23} have specifically used sketching techniques to avoid large entries in the attention matrix. \cite{pmxa23} developed techniques utilizing a transformer within a transformer (TinT) model to simulate the transformer's forward and backward passes, significantly increasing parameter efficiency. \cite{mgn+23} tackled the challenge of fine-tuning LLMs with high memory demands by improving the classical ZO-SCD optimizer, creating a memory-efficient gradient estimator that requires only a forward pass. \cite{bsz23} provided insights into dynamic attention problems, they provide algorithm and hardness for the dynamic setting of attention problem.  \cite{gsyz23} introduces a quantum algorithm for attention computation, opening new avenues for efficiency improvements. \cite{gsy23_dp} provides a result for computing the attention matrix differentially privately. \cite{dms23} introduces a randomized and deterministic attention sparsification algorithms for over-parameterized feature dimension. \cite{dlms23} provides a zero-th order method to accelarate the computation of attention.

\paragraph{Transformer Training}

Transformer architectures (the backbone of LLMs) have been trained with alternating steps of forward and backward computations since their introduction~\cite{vsp+17,dclt18,log+19,ydy+19,bmr+20,zrg+22}. In Appendix~\ref{sec:app_gradient} below, we perform computations to verify that our stated problems are the same as the forward and backward steps from the literature. 

\section{Preliminary}\label{sec:preli}

In Section~\ref{sec:preli:notation}, we define some basic notation we will use. In Section~\ref{sec:preli:mm}, we state important facts related to fast matrix multiplication. In Section~\ref{sec:preli:complexity}, provide the formal definition of the Strong Exponential Time Hypothesis. In Section~\ref{sec:preli:functions}, we define several intermediate functions related to softmax and exponential which will arise in our algorithms. In Section~\ref{sec:preli:loss}, we define the loss function. In Section~\ref{sec:preli:tensor}, we provide standard tensor tricks which we will use. In Section~\ref{sec:preli:reformulate}, we show how to reformulate the loss function for our purposes.

\subsection{Notation}\label{sec:preli:notation}

For any positive integer $n$, we define $[n] := \{1, 2, \dots, n\}$. For two same length vector $x$ and $y$, we use $\langle x, y \rangle$ to denote the inner product between $x$ and $y$, i.e., $\langle x, y \rangle = \sum_{i=1}^n x_i y_i$. We use $x\circ y$ to denote vector that $i$-th entry is $x_i y_i$. Let ${\bf 1}_n$ denote the length-$n$ all ones vector. It is not hard to see that $\langle x \circ y , {\bf 1}_n \rangle = \langle x, y \rangle$. For a vector $x$, we use $x^\top$ to denote the transpose of $x$. For a matrix $M$, we use $M^\top$ to denote the transpose of matrix $M$. For a vector $x$, we use $\exp(z)$ to denote the vector that $i$-th coordinate is $\exp(z_i)$. For a matrix $M$, we use $\exp(M)$ to denote the matrix that $(i,j)$-th coordinate is $\exp(M_{i,j})$. For a function $f$, we use $\wt{O}(f)$ to denote $f \cdot \poly(\log f)$. Let $n_0, n_1, m_0, m_1$ be positive integers. Let $X \in \R^{n_0 \times m_0}$ and $Y \in \R^{n_1 \times m_1}$. We define the Kronecker product between matrices $X$ and $Y$, denoted $X \otimes Y \in \R^{n_0 n_1 \times m_0 m_1}$, as $(X \otimes Y)_{(j_0 - 1) n_1 + j_1, (i_0-1) m_2+i_1}$ 
is equal to $X_{j_0,i_0} Y_{j_1,i_1}$, where $j_0 \in [n_0], i_0 \in [m_0], j_1 \in [n_1], i_1 \in [m_1]$.

\subsection{Matrix Multiplication}\label{sec:preli:mm}
We define matrix multiplication notation and state some well-know facts here.
\begin{definition}
Let $n_1,n_2,n_3$, denote any three positive integers. We use $\Tmat(n_1, n_2, n_3)$ to denote the time of multiplying an $n_1 \times n_2$ matrix with another $n_2 \times n_3$.
\end{definition}

It is well-known that
\begin{fact}[\cite{bcs97,b13}]
Let $n_1,n_2,n_3$, denote any three positive integers. 
$\Tmat(n_1, n_2, n_3) = O(\Tmat(n_1, n_3, n_2)) = O(\Tmat(n_2, n_1, n_3)) = O(\Tmat(n_2,n_3,n_1)) = O( \Tmat(n_3,n_1,n_2) ) = O( \Tmat(n_3,n_2,n_1) )$.
\end{fact}

\subsection{Backgrounds on Complexity}\label{sec:preli:complexity}

Over 20 years ago, Impagliazzo and Paturi \cite{ip01} introduced the Strong Exponential Time Hypothesis ($\SETH$), an enhancement of the $\mathsf{P} \neq \mathsf{NP}$ conjecture. It posits that the existing algorithms for solving $\mathsf{SAT}$ problems are essentially as efficient as possible:

\begin{hypothesis}[Strong Exponential Time Hypothesis ($\SETH$)]\label{hyp:seth}
    For any $\epsilon > 0$, there exists a positive integer $k \geq 3$ for which solving $k$-$\mathsf{SAT}$ problems with $n$ variables in $O(2^{(1-\epsilon )n})$ time is impossible, including with the use of randomized algorithms.
\end{hypothesis}
\noindent{\sf SETH}, a widely recognized conjecture, has been instrumental in establishing fine-grained lower bounds across a broad spectrum of algorithmic challenges, as highlighted in the survey~\cite{w18}.

\subsection{Definitions related with Softmax}\label{sec:preli:functions}

Now, we start by some definitions about $X \in \R^{d \times d}$ which will be helpful. Let $x$ denote the vectorization of $X$.

\begin{definition}\label{def:u}
Let $A_1, A_2 \in \R^{n \times d}$ be two matrices. Suppose that $\A = A_1 \otimes A_2 \in \R^{n^2 \times d^2}$. We define $\A_{j_0} \in \R^{n \times d^2}$ be a $n \times d^2$ size sub-block from $\A$. Note that there $n$ such sub-blocks.

For every $j_0 \in [n]$, let us define function $u(x)_{j_0}: \R^{d^2} \rightarrow \R^n$ to be:
\begin{align*}
    u(x)_{j_0} := \underbrace{ \exp( \A_{j_0} x ) }_{n \times 1}.
\end{align*}
\end{definition}

\begin{definition}\label{def:alpha}
Suppose that there are two $n \times d$ size matrices $A_1, A_2 \in \R^{n \times d}$.  We define $\A_{j_0} \in \R^{n \times d^2}$ be a $n \times d^2$ size sub-block from $\A$.  (Recall that $\A = A_1 \otimes A_2 \in \R^{n^2 \times d^2}$.)

For every index $j_0 \in [n]$, we consider a function, $\alpha(x)_{j_0}: \R^{d^2} \rightarrow \R$ as:
\begin{align*}
  \alpha(x)_{j_0}:= \langle \underbrace{ \exp( \A_{j_0} x ) }_{n \times 1} , \underbrace{ {\bf 1}_n }_{n \times 1} \rangle.
\end{align*}
\end{definition}

\begin{definition}\label{def:f}

Suppose that $\alpha(x)_{j_0} \in \R$ is defined as in Definition~\ref{def:alpha}.

Recall $u(x)_{j_0} \in \R^n$ is defined as in Definition~\ref{def:u}.

For a fixed $j_0 \in [n]$, let us consider function $f(x)_{j_0} : \R^{d^2} \rightarrow \R^n$
\begin{align*}
    f(x)_{j_0} := \underbrace{ \alpha(x)_{j_0}^{-1} }_{ \mathrm{scalar} } \underbrace{ u(x)_{j_0} }_{ n \times 1 } .
\end{align*}
Let $f(x) \in \R^{n \times n}$ denote the matrix where $j_0$-th row is $( f(x)_{j_0} )^\top$.
\end{definition}

\begin{definition}\label{def:h}
For every $i_0 \in [d]$, we define $h()_{i_0} : \R^{d^2} \rightarrow \R^n$ as:
\begin{align*}
    h(y)_{i_0}:= \underbrace{ A_3 }_{n \times d} \underbrace{ Y_{*,i_0} }_{d \times 1}.
\end{align*}
Here let $Y \in \R^{d \times d}$ denote the matrix representation of $y \in \R^{d^2}$. Let $h(y) \in \R^{n \times d}$ matrix where $i_0$ column is $h(y)_{i_0}$.
\end{definition}

\subsection{Loss Functions}\label{sec:preli:loss}

In this section, we introduce some helpful definitions related to both $x \in \R^{d^2}$.

\begin{definition}\label{def:c}
For every $j_0 \in [n]$, we use $f(x)_{j_0} \in \R^n$ to denote the normalized vector defined by Definition~\ref{def:f}. For every $i_0 \in [d]$, we let $h(y)_{i_0}$ to be defined in Definition~\ref{def:h}.

Consider every $j_0 \in [n]$, every $i_0 \in [d]$. Let us consider $c(x)_{j_0,i_0}: \R^{d^2} \times \R^{d^2} \rightarrow \R$ as follows:
\begin{align*}
    c(x)_{j_0,i_0}:= \langle f(x)_{j_0}, h(y)_{i_0} \rangle - E_{j_0,i_0}.
\end{align*}
Here $E_{j_0,i_0}$ is the $(j_0,i_0)$-th coordinate/location of $E \in \R^{n \times d}$ for $j_0 \in [n], i_0 \in [d]$.
This is equivalent to $\underbrace{ c(x) }_{n \times d} = \underbrace{ f(x) }_{n \times n} \underbrace{ h(y) }_{n \times d} - \underbrace{E}_{n \times d}$.
\end{definition}

\begin{definition}\label{def:l}
For every $j_0 \in [n]$, for every $i_0 \in [d]$. Let us define
$
 L(x)_{j_0,i_0} $ to  be $:= 0.5 c(x)_{j_0,i_0}^2
 $. 
\end{definition}

\subsection{Tensor Trick}\label{sec:preli:tensor}

We state the well-known tensor-trick. It has been widely used in literature of linear algebra related to tensor computations \cite{swz19_soda,dssw18,djs+19,as24,gsx23_incontext,z22,rsz22,gsy23_coin, dsy23, dgs23}.
\begin{fact}[Tensor trick]\label{fac:tensor_trick_basic}
For two matrices $A_1$ and $A_2 \in \R^{n \times d}$, define $\A = A_1 \otimes A_2$. Let $X \in \R^{d \times d}$. Let $x \in \R^{d^2}$ denote the vector representation of $X$.  Then we have
    $\vect ( A_1 X A_2^\top ) = \A x$.
\end{fact}

Using the above tensor-trick, it is easy to observe that
\begin{fact}\label{fac:tensor_trick_more}
For two matrices $A_1$ and $A_2 \in \R^{n \times d}$, denote $\A = A_1 \otimes A_2$. Let $X \in \R^{d \times d}$. Let $\A_{j_0} \in \R^{n \times d^2}$ a submatrix of $\A$ (by properly selecting $n$ rows of $\A$). Let $x \in \R^{d^2}$ denote the vector representation of $X$.
Then, we have
\begin{itemize}
    \item $\vect ( \exp( A_1 X A_2^\top ) ) = \exp( \A x)$
    \item $ ( \exp(A_1 X A_2^\top)_{j_0,*} )^\top = \exp( \A_{j_0} x ) $, 
\end{itemize}
Here $\exp(A_1 X A_2^\top)_{j_0,*}$ is the $j_0$-th row of $n \times n$ matrix $\exp(A_1 X A_2^\top)$.
\end{fact}
\begin{proof}
We can use the definition in Lemma and Definition~\ref{fac:tensor_trick_basic}, to prove it.
\end{proof}

\subsection{Reshape the Loss function via Tensor Trick}\label{sec:preli:reformulate}

\begin{lemma}\label{lemma:L}
Given the below requirements
\begin{itemize}
    \item Here are three matrices $A_1 \in \R^{n \times d}$, $A_2 \in \R^{n \times d}$, and $A_3 \in \R^{n \times d}$
    \item Let $\mathsf{A} = A_1 \otimes A_2 \in \R^{n^2 \times d^2}$ to be the Kronecker product of the two matrices $A_1$ and $A_2$
    \begin{itemize}
        \item For every $j_0 \in [n]$, define $\A_{j_0} \in \R^{n \times d^2}$ to be a $n \times d^2$ sized block in the matrix $\A \in \R^{n^2 \times d^2}$
    \end{itemize} 
    \item $E \in \R^{n \times d}$ be a matrix. Define $E_{j_0,i_0}$ as the $(j_0,i_0)$-th coordinate/location of $E \in \R^{n \times d}$ for every pair of $j_0 \in [n]$ and $i_0 \in [d]$  
    \item Here are two square matrices $X \in \R^{d \times d}$, let $Y \in \R^{d \times d}$
    \item Let $L(X)$ be defined as Definition~\ref{def:attention_optimization_loss}
    \item For every pair of $j_0 \in [n]$, $i_0 \in [d]$, recall that definition of $L(x)_{j_0,i_0}$ can be found in in Definition~\ref{def:l}
\end{itemize}
Then, we have
\begin{align*}
    L(X) = \sum_{j_0 \in [n]} \sum_{i_0 \in [d]} L(x)_{j_0,i_0}.
\end{align*}
\end{lemma}
\begin{proof}
We can show that
\begin{align*}
    & ~ L(X) \\
    = & ~0.5 \cdot \| \underbrace{ D(X)^{-1} }_{n \times n} \underbrace{ \exp(A_1 X A_2^\top) }_{n \times n} \underbrace{ A_3 }_{n \times d} \underbrace{Y}_{d \times d} - \underbrace{ E  }_{n \times d} \|_F^2 \\
    = & ~ \sum_{j_0=1}^n \sum_{i_0=1}^d 0.5 \cdot\\
    & ~ ( \langle \langle \exp(\A_{j_0} x ) , {\bf 1}_n \rangle^{-1} \cdot \exp (\A_{j_0} x ) , A_3 Y_{*,i_0} \rangle - E_{j_0,i_0})^2  \\
    = & ~ \sum_{j_0=1}^n \sum_{i_0=1}^d 0.5 ( \langle f(x)_{j_0}, h(y)_{i_0} \rangle - E_{j_0,i_0} )^2 \\
    = & ~ \sum_{j_0=1}^n \sum_{i_0=1}^d L(x)_{j_0,i_0}
\end{align*}
where the first step follows from definition, the second step follows from writing down the summation, the third step follows from definition of $f(x)_{j_0}$ (recall the Definition~\ref{def:f}) and $h(y)_{i_0}$ (recall the Definition~\ref{def:h}), and the last step follows from $L(x)_{j_0,i_0}$ (see Definition~\ref{def:l}).
\end{proof}

\section{Proof Sketch for General Upper Bound}\label{sec:upper_bound}

The most straightforward way to compute the gradient would take $O(n^2 d^2)$ time in order to explicitly write down the matrix $\A$. By using fast matrix multiplication and regroup the entries, we can obtain our first intermediate algorithm, which runs in quadratic time to compute the gradient.
\begin{lemma}[Attention gradient computation, informal version of Lemma~\ref{lem:gradient:formal}]\label{lem:gradient:informal}

If the following conditions hold
\begin{itemize}
    \item Define four $n \times d$ size matrices $E, A_1, A_2, A_3$ and two $d \times d$ square matrices $X, Y$ to be input fixed matrices.
    \item Let $X \in \R^{d \times d}$ and $Y \in \R^{d \times d}$ denote matrix variables (we will compute gradient with respect to $X$ )
    \begin{itemize}
        \item For easy of writing, we also use vector variables $x \in \R^{d^2 \times 1}$ and $y \in \R^{d^2 \times 1}$
    \end{itemize}
    \item Let $g = \frac{\d L(X)}{\d x} \in \R^{d^2}$ (We abuse notation $L(x)$ and $L(X)$ are the same thin)
\end{itemize}
Then we can show that gradient $g \in \R^{d^2}$ can be calculated in $O(\Tmat(n,d,n) + \Tmat(n,d,d))$ time.
\end{lemma}

Next, we will show how to improve the running time of computing gradient from quadratic time ($\geq n^2$) to almost linear time $n^{1+o(1)}$.

Note that by linearity of derivative, we can show that
\begin{align*}
    \frac{ \d L(x) }{\d x } = \sum_{j_0=1}^n \sum_{i_0=1}^d \frac{\d L(x)_{j_0,i_0} }{ \d x } 
\end{align*}

Based on calculations we perform in Section~\ref{sec:app_gradient}, Section~\ref{sec:app_time}, and several linear algebra facts, we can show that
\begin{align*}
    & ~\frac{\d L(x)_{j_0,i_0} }{ \d x } \\
    = & ~ \underbrace{ c(x)_{j_0,i_0} }_{ \mathrm{scalar} } \cdot \underbrace{ \A_{j_0}^\top }_{d^2 \times n} \underbrace{ ( \diag( f(x)_{j_0} ) - f(x)_{j_0} f(x)_{j_0}^\top ) }_{n \times n} \underbrace{ h(y)_{i_0} }_{n \times 1}
\end{align*}

For any fixed $j_0 \in [n]$, consider this quantity. Since this expression involves an $n\times n$ matrix, the most straightforward way to calculate it would take $\Theta(n^2)$ time, and so summing over all $j_0 \in [n]$ would lead to a cubic-time algorithm. It is not too difficult to improve this: the $n \times n$ matrix
\begin{align*}
( \underbrace{ \diag( f(x)_{j_0} ) }_{\mathrm{a~diagonal~matrix}} - \underbrace{ f(x)_{j_0} f(x)_{j_0}^\top }_{\mathrm{a~rank~1~matrix}} )
\end{align*}
is easily decomposed into a low-rank part ($ f(x)_{j_0} f(x)_{j_0}^\top $ which has size $n \times n$) and a sparse part ($\diag( f(x)_{j_0} )  $ which also has size $n \times n$), which reduces the calculation of each part to only $\tilde{O}(n)$ time, and the total running time to $\tilde{O}(n^2)$ time.

However, we are aiming for a almost-linear time algorithm, and it is not possible to achieve this by treating the different $j_0$ separately, since a given $j_0$ must take $\Omega(n)$ time to process. Instead, we use tensor techniques related to low-rank approximations to simultanouesly compute all $j_0$ together and sum them in almost-linear time.

In order to do that, we create several extra artificial or intermediate matrices $q(x) \in \R^{n \times n}$(see Section~\ref{sec:app_time}), $p(x) \in \R^{n \times n}$ (see Section~\ref{sec:app_time}). We will show the gradient can be finally constructed using a simple chaining technique (see Section~\ref{sec:app_fast_time} for more details), from $f, c, q$, $p_1$ (handling $\diag( f(x)_{j_0})$ similarly), $p_2$ (handling $f(x)_{j_0} f(x)_{j_0}^\top$ similarly), $p$ ($p= p_1- p_2$) to $\frac{\d L}{ \d x}$. Intuitively, the chaining shows that a low rank representation for $f$ yields one for $c$, and these in turn yield one for $q$, and so on.

In particular, using $q(x)$, we obtain that $\frac{\d L(x)}{\d x}$ can be written as
\begin{align*}
\sum_{j_0=1}^n \A_{j_0}^\top ( \underbrace{ \mathrm{~a~diagonal~matrix}}_{\diag( f(x)_{j_0} )} - \underbrace{ \mathrm{~a~rank~1~matrix} }_{ f(x)_{j_0} f(x)_{j_0}^\top } ) \underbrace{ \mathrm{a~column~vector}}_{ q(x)_{j_0} }  
\end{align*}
which in fact notably removes the summation step of $i_0=1$ to $d$. Using the notation of $p(x)$, we finally yield that we need to compute $A_1^\top p(x) A_2$. Thus as long as $p(x)$ has a low-rank representation, then we can solve the in $n^{1+o(1)}$ time (see Section~\ref{sec:app_fast_time} for more details). In particular, we will find that $p(x)$ is the entry-wise product of two matrices with low-rank representations from prior work, which we can combine using a column-wise Kronecker product to approximate $p(x)$ itself.

\section{General Lower Bound}\label{sec:lower_bound}

We will critically make use of the known hardness result for attention computation itself, which we state now.

\begin{definition}[Attention Computation] \label{attcomp}
    Given as input matrices $Q,K,V \in \R^{n \times d}$ and a parameter $\varepsilon > 0$, compute a matrix $T \in \R^{n \times d}$ satisfying $$\|T - D^{-1} A V\|_\infty \leq \varepsilon,$$
    where $A = \exp(QK^\top)$ and $D = \diag(A {\bf 1}_n)$.
\end{definition}

\begin{lemma}[Lemma 4.7 in \cite{as23}] \label{prevlemma}
    Assuming $\SETH$, there is no algorithm running in time $O(n^{2 - \delta})$ for any constant $\delta>0$ that solves Attention Computation (Definition~\ref{attcomp}), even when the inputs satisfy the following constraints, for any parameter $\kappa \geq 0$:
    \begin{itemize}
        \item $d = O(\log n)$,
        \item $V \in \{0,1\}^{n \times d}$,
        \item There is a value $B \leq O(\log^2 n \cdot (1+\kappa))$ such that every entry of $QK^\top$ is in the interval $[0,B]$ and at least half the entries in each row of $QK^\top$ are equal to $B$,
        \item moreover $\|Q\|_\infty, \|K\|_\infty \leq O(\sqrt{\log n (1 + \kappa)})$, and
        \item $\varepsilon < n^{\kappa - O(1)}$.
    \end{itemize}
\end{lemma}

Next, we show that the attention optimization problem behaves particularly well when given matrices constrained as in Lemma~\ref{prevlemma}:

\begin{lemma} \label{boundedderivs}
    Let $A$ be a fixed $n \times n$ matrix whose entries are real numbers in the interval $[0,B]$, and such that in each row of $A$, at least half the entries are equal to $B$. Let $V$ be any $n \times d$ matrix whose entries are all in $\{0,1\}$. For $\lambda \in \R$, define the $n \times n$ matrix $M_\lambda := \exp(\lambda A)$, where $\exp$ is applied entry-wise. Define the function $f : \R \to \R$ by
    \begin{align*}
        f(\lambda) := \| \diag(M_\lambda {\bf 1}_n)^{-1} M_\lambda V \|_F^2,
    \end{align*}
    Then, for all $\lambda \in \R$ we have
    \begin{itemize}
        \item $|f'(\lambda)| \leq O(Bn)$, 
        \item $|f''(\lambda)| \leq O(B^2 n)$.
    \end{itemize}
\end{lemma}

\begin{proof}
Let $C$ denote the $n \times n$ matrix $C =  \diag(M_\lambda {\bf 1}_n)^{-1} M_\lambda$. For $i,j \in [n]$, we  calculate that $M_{\lambda}[i,j] = e^{\lambda A[i,j]}$ and so
\begin{align*}
C[i,j] = \frac{e^{\lambda A[i,j]}}{\sum_{k=1}^n e^{\lambda A[i,k]}}.
\end{align*}

For $\ell \in [d]$, let $S_\ell \subseteq [n]$ be the set of $1$s in column $\ell$ of $V$, i.e., $S_\ell = \{ j \in [n] \mid V[j,\ell] = 1\}$. Hence, for $i \in [n]$ and $\ell \in [d]$, the entry $(i,\ell)$ of the matrix $\diag(M_\lambda {\bf 1}_n)^{-1} M_\lambda V$ is given by
\begin{align*}
    \diag(M_\lambda {\bf 1}_n)^{-1} M_\lambda V [i,\ell] &= CV[i,\ell] 
    \\ &= \sum_{j=1}^n C[i,j] V[j,\ell]
    \\ &= \sum_{j \in S_\ell} C[i,j]
    \\ &= \frac{\sum_{j \in S_\ell} e^{\lambda A[i,j]}}{\sum_{k=1}^n e^{\lambda A[i,k]}}.
\end{align*}
where the first step follows from definition, the second step follows from simple algebra.

We thus get an explicit expression for $f(\lambda)$:

\begin{align*}
f(\lambda) &= \sum_{i = 1}^n \frac{\sum_{\ell=1}^d \left(\sum_{j \in S_\ell} e^{\lambda A[i,j]} \right)^2}{\left(\sum_{k=1}^n e^{\lambda A[i,k]}\right)^2}
\\ &= \sum_{i = 1}^n \frac{\sum_{\ell=1}^d \sum_{j_1 \in S_\ell}^n \sum_{j_2 \in S_\ell}^n e^{\lambda (A[i,j_1] + A[i,j_2])}}{\sum_{k_1=1}^n \sum_{k_2=1}^n e^{\lambda (A[i,k_1] + A[i,k_2])}}.
\end{align*}

We define 
\begin{align*}
a(\lambda,i) := \sum_{\ell=1}^d \sum_{j_1 \in S_\ell}^n \sum_{j_2 \in S_\ell}^n e^{\lambda (A[i,j_1] + A[i,j_2])}
\end{align*}
and then we define
\begin{align*}
b(\lambda,i) := \sum_{k_1=1}^n \sum_{k_2=1}^n e^{\lambda (A[i,k_1] + A[i,k_2])}
\end{align*}
Combining the above three equations, we can obtain 
\begin{align*}
    f(\lambda) = \sum_{i=1}^n a(\lambda,i)/b(\lambda,i).
\end{align*}

Since, for each row of $A$, at least half the entries equal $B$, and all the entries are in the interval $[1,B]$, we can bound
\begin{align}\label{eq:bound_on_b_lambda_i}
\left( \frac{n}{2} \right)^2 \cdot e^{2B\lambda} \leq b(\lambda,i) \leq \left( n \right)^2 \cdot e^{2B\lambda}.
\end{align}

Furthermore, since the derivative of $e^{\lambda (A[i,k_1] + A[i,k_2])}$ with respect to $\lambda$ is $(A[i,k_1] + A[i,k_2]) \cdot e^{\lambda (A[i,k_1] + A[i,k_2])}$, we can bound
\begin{align}\label{eq:bound_on_d_b_lambda_i}
2 \cdot b(\lambda,i) \leq \frac{\d b(\lambda,i)}{\d \lambda} \leq 2B \cdot b(\lambda,i).
\end{align}

We may similarly bound
\begin{align}\label{eq:bound_on_a_lambda_i}
0 \leq a(\lambda,i) \leq n^2 \cdot e^{2B\lambda},
\end{align}
and
\begin{align}\label{eq:bound_on_d_a_lambda_i}
2 \cdot a(\lambda,i) \leq \frac{\d a(\lambda,i)}{\d \lambda} \leq 2B \cdot a(\lambda,i).
\end{align}

We can thus bound the derivative of $f$ (where here, all the $'$ notation means derivative with respect to $\lambda$):
\begin{align*}
    f'(\lambda) &= \sum_{i=1}^n \frac{a'(\lambda,i) \cdot b(\lambda,i) - a(\lambda,i) \cdot b'(\lambda,i)}{(b(\lambda,i))^2}
    \\ &\leq \sum_{i=1}^n \frac{a'(\lambda,i) \cdot b(\lambda,i) }{(b(\lambda,i))^2}
    \\ &= \sum_{i=1}^n \frac{a'(\lambda,i)  }{b(\lambda,i)}
    \\ &\leq \sum_{i=1}^n \frac{2B \cdot n^2 e^{2B\lambda}}{(n/2)^2 \cdot e^{2B\lambda}}
    \\ &= \sum_{i=1}^n 8B
    \\ &= 8B \cdot n.
\end{align*}
where the 1st step follows from definition, the 2nd step follows from simple algebra, the 3rd step follows from cancelling $b(\lambda,i)$, the 4th step is using  Eq.~\eqref{eq:bound_on_b_lambda_i} (for $b(\lambda,i)$) and Eq.~\eqref{eq:bound_on_d_a_lambda_i} (for $a'(\lambda,i)$), the 5th step follows from simple algebra, and the last step follows from simple algebra.

Similarly, we can provide a lower bound $f'(\lambda)$,
\begin{align*}
    f'(\lambda) &= \sum_{i=1}^n \frac{a'(\lambda,i) \cdot b(\lambda,i) - a(\lambda,i) \cdot b'(\lambda,i)}{(b(\lambda,i))^2}
    \\ &\geq - \sum_{i=1}^n \frac{a(\lambda,i) \cdot b'(\lambda,i)}{(b(\lambda,i))^2}
    \\ &\geq - \sum_{i=1}^n \frac{(n^2 \cdot e^{2B\lambda}) \cdot (2B \cdot b(\lambda,i))}{((n/2)^2 \cdot e^{2B\lambda}) \cdot (b(\lambda,i))}
    \\ &= - \sum_{i=1}^n 8B
    \\ &= - 8B \cdot n.
\end{align*}
where the 1st step follows from definition, the 2nd step follows form simple algebra, the 3rd step follows Eq.~\eqref{eq:bound_on_d_b_lambda_i} (for $b'(\lambda,i)$) and Eq.~\eqref{eq:bound_on_a_lambda_i} (for $a(\lambda,i)$), the 4th step follows from simple algebra, and the last step follows from simple algbera.

Finally, letting $f(\lambda,i) := a(\lambda,i)/b(\lambda,i)$, we have again by the quotient rule that $f''(\lambda)$ is equal to
$$\sum_{i=1}^n \frac{a''(\lambda,i)  - b''(\lambda,i) \cdot f(\lambda,i) - 2 \cdot b'(\lambda,i) \cdot f'(\lambda,i)}{b(\lambda,i)}$$

which we similarly bound in magnitude by $O(B^2 n)$.
\end{proof}

We recall a simple approximation from calculus:

\begin{lemma} \label{calculus}
    Let $f : [0,1] \to \R$ be a twice-differentiable function such that $|f''(\lambda)| \leq b$ for all $\lambda \in [0,1]$. For any positive integer $m$, define the sum
    $$t_m := \sum_{i=0}^{m-1} \frac{f'(i/m)}{m}.$$
    Then,
    $$|t_m - (f(1) - f(0))| \leq b/m.$$
\end{lemma}

\begin{proof}
    If two $\lambda_0, \lambda_1 \in [0,1]$ have $|\lambda_0 - \lambda_1| \leq 1/m$, then from our bound on $f''(\lambda)$, we know that $|f'(\lambda_1) - f'(\lambda_0)| \leq b/m$.
    We can thus bound the difference
    \begin{align*}
    f(1) - f(0) = \int_0^1 f'(\lambda) d\lambda
    \end{align*}
    by
    \begin{align*}
    f(1) - f(0) \leq \sum_{i=0}^{m-1} \frac{f'(i/m) + (b/m)}{m} = t_m + b/m
    \end{align*}
    and
    \begin{align*}
        f(1) - f(0) \geq \sum_{i=0}^{m-1} \frac{f'(i/m) - (b/m)}{m} = t_m - b/m.
    \end{align*}
    Thus, we complete the proof.
\end{proof}

Finally, we are ready for our main result:

\begin{theorem}[Formal version of Theorem~\ref{thm:mainlb}]\label{thm:mainlb:formal}
    Let $\kappa : \N \to \N$ by any function with $\kappa(n) = \omega(1)$ and $\kappa(n) = o(\log n)$.
    Assuming $\SETH$,  there is no algorithm running in time $O(n^{2 - \delta})$ for any constant $\delta>0$ for Approximate Attention Loss Gradient Computation (Definition~\ref{def:AAttLGC}), even in the case where $d = O(\log n)$ and the input matrices satisfy $\|A_1\|_\infty, \|A_2\|_\infty, \|A_3\|_\infty \leq O(\sqrt{\log n} \cdot \kappa(n))$, $B = 0$, $Y = I$, $X = \lambda I$ for some scalar $\lambda \in [0,1]$, and $\varepsilon = O(1/(\log n)^4)$.
\end{theorem}

\begin{proof}
    Suppose there were such an algorithm. We call it $O((\log n)^4)$ times to refute Lemma~\ref{prevlemma} (with parameter $\kappa = \kappa(n)$). Let $Q,K,V$ be the input matrices to Lemma~\ref{prevlemma}, and set $A_1 = Q$, $A_2 = K$, $A_3 = V$, $Y= I$, and $X = \lambda I$ for a parameter $\lambda \in [0,1]$. Suppose the function $f : [0,1] \to \R$ is in Lemma~\ref{boundedderivs} where $A$ is the matrix $A_1 A_2^\top$, so that $M_\lambda$ is the matrix $\exp(A_1 X A_2^\top)$. It follows from Lemma~\ref{boundedderivs} that 
    \begin{align*}
        |f''(\lambda)| \leq O(n \log^2 n \cdot (\kappa(n))^2).
    \end{align*}

    We can compute $f(0)$ in $\tilde{O}(n)$ time since then $M_f$ is the all-1s matrix, and our goal is to output $f(1)$. 
    
    Thus, by Lemma~\ref{calculus}, it suffices to compute $f'(\lambda)$ on 
    $
        O(\log^2(n) (\kappa(n))^2) = O(\log^4 n)
    $
    points up to $O(1/(\log n)^4)$ error, and return their average. But, since we have picked $X = \lambda I$, we can calculate $f'(\lambda)$ from the gradient $\frac{\d L(X)}{\d X}$ (from Definition~\ref{def:AAttLGC}), which is approximated by our assumed algorithm.
\end{proof}

\section{Conclusion}\label{sec:conclusion}

Our results give a complete fine-grained analysis of the running time needed to train LLMs. We show that there is a threshold depending on the parameter $B$, the magnitude of the parameter matrix entries. In settings where $B$ is small, a near-linear-time algorithm for LLM training is possible by using our novel algorithm for backward computation. In settings where $B$ is large, not only does our algorithm not apply, but we show it is impossible to design a nontrivially-fast algorithm (barring a breakthrough in satisfiability algorithms that would refute the popular $\SETH$).

These insights can guide LLM designers to more efficient algorithms. When $B$ can be made small, it would lead to substantial savings in the computational resources needed for training and expression. When $B$ must be large (perhaps to achieve a high expressiveness?), our lower bounds show that one may as well use straigthforward algorithms and focus on other aspects of algorithm speedup such as parallelization. The magnitude of $B$ needed has been studied more recently (e.g.,~\cite{as24}), and the need for fast training algorithms may further motivate this direction of research.

\ifdefined\isarxiv

\else
\section*{Impact Statement}

This paper addresses computational concerns for implementing known algorithms from the field of Machine Learning. There are many potential societal consequences of our work, none which we feel must be specifically highlighted here.
\fi

\ifdefined\isarxiv 
\else
\bibliography{ref}
\bibliographystyle{icml2024}

\fi

\newpage
\onecolumn
\appendix
\section*{Appendix}

{\bf Roadmap.}

In Section~\ref{sec:app_preli}, we provide basic notation and facts. In Section~\ref{sec:app_gradient}, we provide details about gradient computations. In Section~\ref{sec:app_time}, we explain the computation time for the gradient of attention loss. In Section~\ref{sec:app_fast_time}, we show how to further improve the gradient computation from quadratic time to almost linear time.

\section{Preliminaries}\label{sec:app_preli}
In Section~\ref{sec:app_preli:notations}, we define some basic notation. In Section~\ref{sec:app_preli:facts}, we state several facts which we will use.

\subsection{Notation}\label{sec:app_preli:notations}

For any positive integer $n$, we define $[n] := \{1, 2, \dots, n\}$.

For two same length vector $x$ and $y$, we use $\langle x, y \rangle$ to denote the inner product between $x$ and $y$, i.e., $\langle x, y \rangle = \sum_{i=1}^n x_i y_i$. We use $x\circ y$ to denote vector that $i$-th entry is $x_i y_i$. Let ${\bf 1}_n$ denote the length-$n$ all ones vector. It is not hard to see that $\langle x \circ y , {\bf 1}_n \rangle = \langle x, y \rangle$.

For a vector $u$, we use $u^\top$ to denote the transpose of $u$. For a matrix $M$, we use $M^\top$ to denote the transpose of matrix $M$.

For a vector $u$, we use $\exp(u)$ to denote the vector that $i$-th coordinate is $\exp(u_i)$. For a matrix $A$, we use $\exp(A)$ to denote the matrix that $(i,j)$-th coordinate is $\exp(A_{i,j})$.

We define the Kronecker product between matrices $X$ and $Y$, denoted $X \otimes Y \in \R^{n_0 n_1 \times m_0 m_1}$, as $(X \otimes Y)_{(j_0 - 1) n_1 + j_1, (i_0-1) m_2+i_1}$ 
is equal to $X_{j_0,i_0} Y_{j_1,i_1}$, where $j_0 \in [n_0], i_0 \in [m_0], j_1 \in [n_1], i_1 \in [m_1]$.

For each positive integers $m_1,m_2,m_3$, we use $\Tmat(m_1,m_2,m_3)$ to denote the time of multiplying $m_1 \times m_2$ matrix with another $m_2 \times m_3$ matrix.

\subsection{Basic Facts}\label{sec:app_preli:facts}

\begin{fact}\label{fac:circ_rules}
Let $x,y,z \in \R^n$. Then we have
\begin{itemize}
    \item $\langle x \circ y, z\rangle = x^\top \diag(y) z$.
    \item $\langle x, y \rangle = \langle x \circ y, {\bf 1}_n \rangle$.
\end{itemize}
\end{fact}

\begin{fact}[Folklore]\label{fac:olash_folklore}
Let $U_1, V_1 \in \R^{n \times k_1}$. Let $U_2, V_2 \in \R^{n \times k_2}$. Then we have
\begin{align*}
    (U_1 V_1^\top) \circ (U_2 V_2^\top) = (U_1 \oslash U_2)  (V_1 \oslash V_2)^\top
\end{align*}

Here, given $U_1 \in \R^{n \times k_1}$ and $U_2 \in \R^{n \times k_2}$, the $U_1 \oslash U_2 \in \R^{n \times k_1 k_2}$ is the row-wise Kronecker product, i.e., $(U_1 \oslash U_2)_{i, l_1 + (l_2 - 1)k_1 }:= (U_1)_{i,l_1} U_{i,l_2}$ for all $i \in [n]$, $l_1 \in [k_1]$ and $l_2 \in [k_2]$
\end{fact}

\section{More Details about Gradient Computation}\label{sec:app_gradient}

In this section, we provide details and calculations to assist with gradient and derivative computations. We remark that, in this section, for convenience of computing a closed form for the gradient, we ignore the $1/d$ factor in function $f$. Since it is only a rescaling factor, it won't affect how we compute these matrices in general.

\begin{lemma}[The gradient computation for several different functions with respect to $x_i$]\label{lem:gradient_x}
For every $i \in [d^2]$, define $\A_{j_0,i} \in \R^n$ to be the $i$-th column for $\A_{j_0} \in \R^{n \times d}$. $u(x)_{j_0} \in \R^n$. The scalar function $\alpha(x)_{j_0} \in \R$, column function $f(x)_{j_0} \in \R^n$, scalar function $c(x)_{j_0,i_0} \in \R$ and  scalar function $L(x)_{j_0,i_0} \in \R$ are defined as in Definitions~\ref{def:u}, \ref{def:alpha}, \ref{def:f}, \ref{def:c} and \ref{def:l} respectively. 
 
Then, for each $i \in [d^2]$, we have
\begin{itemize}
    \item {\bf Part 1.}
    \begin{align*}
        \frac{ \d x }{\d x_i} = e_i
    \end{align*}
    \item {\bf Part 2.} For each $j_0 \in [n]$,
    \begin{align*}
        \frac{ \d \A_{j_0} x }{ \d x_i } = (\A_{j_0})_{i}
    \end{align*}
    \item {\bf Part 3.} For each $j_0 \in [n]$
    \begin{align*}
        \frac{ \d u(x)_{j_0} }{ \d x_i } = \A_{j_0,i} \circ u(x)_{j_0}
    \end{align*}
    \item {\bf Part 4.} For each $j_0 \in [n]$,  
    \begin{align*}
        \frac{\d \alpha(x)_{j_0}}{\d x_i} = \langle \A_{j_0,i}, u(x)_{j_0} \rangle
    \end{align*}
    \item {\bf Part 5.} For each $j_0 \in [n]$, 
    \begin{align*}
        \frac{ \d f(x)_{j_0} }{ \d x_i } = \A_{j_0,i} \circ f(x)_{j_0}  - \langle \A_{j_0,i}, f(x)_{j_0}\rangle \cdot f(x)_{j_0}
    \end{align*}
    \item {\bf Part 6.} For each $j_0 \in [n]$, for each $i_0 \in [d]$, 
    \begin{align*}
        \frac{ \d \langle f(x)_{j_0} , h(y)_{i_0} \rangle }{ \d x_i } = \langle h(y)_{i_0}, \A_{j_0,i} \circ f(x)_{j_0} \rangle - \langle  h(y)_{i_0}, f(x)_{j_0} \rangle \cdot \langle \A_{j_0,i}, f(x)_{j_0} \rangle
    \end{align*} 
    \item {\bf Part 7.} For each $j_0 \in [n]$, for every $i_0 \in [d]$
    \begin{align*}
        \frac{ \d c(x)_{j_0,i_0} }{ \d x_i } = \langle \A_{j_0,i} \circ  f(x)_{j_0}, h(y)_{i_0} \rangle - \langle  f(x)_{j_0} , h(y)_{i_0} \rangle \cdot \langle\A_{j_0,i}, f(x)_{j_0} \rangle
    \end{align*}
    \item {\bf Part 8.} For each $j_0 \in [n]$, for each $i_0 \in [d]$
    \begin{align*}
        \frac{\d L(x)_{j_0,i_0}}{\d x_i} = (\langle h(y)_{i_0} , \A_{j_0,i} \circ f(x)_{j_0}\rangle - \langle f(x)_{j_0}, \A_{j_0,i} \rangle \cdot \langle h(y)_{i_0} , f(x)_{j_0}\rangle) \cdot c(x)_{j_0, i_0}
    \end{align*}
\end{itemize}
\end{lemma}
\begin{proof}

{\bf Proof of Part 1.}
We have
\begin{align*}
    \frac{\d x }{\d x_i}
\end{align*}

{\bf Proof of Part 2.}
We have
\begin{align*}
    \frac{\d \A_{j_0} x }{ \d x_i} = & ~ \underbrace{ \A_{j_0} }_{n \times d^2} \underbrace{ \frac{\d x}{ \d x_i} }_{d^2 \times 1} \\
    = & ~ \underbrace{ \A_{j_0} }_{n \times d^2} \cdot \underbrace{ e_i }_{d^2 \times 1} \\
    = & ~ \A_{j_0,i}
\end{align*}

{\bf Proof of Part 3.}

We can show

\begin{align*}
    \frac{ \d u(x)_{j_0} }{ \d x_i } 
    = & ~ \frac{ \d \exp( \A_{j_0} x ) }{ \d x_i } \\
    = & ~ \exp( \A_{j_0} x ) \circ \frac{\d \A_{j_0} x }{ \d x_i} \\
    = & ~  \exp( \A_{j_0} x ) \circ \A_{j_0,i} \\
    = & ~ u(x)_{j_0} \circ \A_{j_0,i} 
\end{align*}
where the 3rd step follows from Part 2, the last step follows from definition of $u(x)_{j_0}$.

{\bf Proof of Part 4.}

 For simplicity of writing proofs, we use $(\cdot)$ to denote $(x)$.
 
We can show
\begin{align*}
 \frac{ \d \alpha(\cdot)_{j_0} }{\d x_i} 
 = & ~ \frac{\d \langle u(\cdot)_{j_0} , {\bf 1}_n \rangle }{\d x_i} \\
 = & ~ \langle u(\cdot)_{j_0} \circ \A_{j_0,i} , {\bf 1}_n \rangle \\
 = & ~ \langle u(\cdot)_{j_0} , \A_{j_0,i} \rangle
\end{align*}
where the 1st step follows from definition of $\alpha(\cdot)$, the 2nd step follows from Part 3, the 3rd step follows from Fact~\ref{fac:circ_rules}.

{\bf Proof of Part 5.}
 For simplicity of writing proofs, we use $(\cdot)$ to denote $(x)$.

We can show that 
\begin{align*}
        \frac{ \d f(\cdot)_{j_0} }{ \d x_i } 
        = & ~  \frac{ \d \alpha(\cdot)_{j_0}^{-1} u(\cdot)_{j_0} }{ \d x_i } \\
        = & ~ \alpha(\cdot)_{j_0}^{-1} \frac{\d u(\cdot)_{j_0}}{\d x_i} + ( \frac{\d \alpha(\cdot)_{j_0}^{-1} }{\d x_i} ) u(\cdot)_{j_0} 
\end{align*}
For the first term, we have
\begin{align*}
    \alpha(\cdot)_{j_0}^{-1} \frac{\d u(\cdot)_{j_0}}{\d x_i} 
    = & ~ \alpha(\cdot)_{j_0}^{-1} u(\cdot)_{j_0} \circ \A_{j_0,i} \\
    = & ~ f(\cdot)_{j_0} \circ \A_{j_0,i}
\end{align*}
where the 1st step follows from Part 3, the 2nd step follows from definition of $f(\cdot)$.

For the second term, we have
\begin{align*}
    ( \frac{\d \alpha(\cdot)_{j_0}^{-1} }{\d x_i} ) u(\cdot)_{j_0} 
    = & ~ - \alpha(\cdot)_{j_0}^{-2} \frac{ \d \alpha(\cdot)_{j_0} }{\d x_i} u(\cdot)_{j_0} \\
    = & ~ - \alpha(\cdot)_{j_0}^{-2} \cdot \langle u(\cdot)_{j_0} , \A_{j_0,i} \rangle \cdot u(\cdot)_{j_0} \\
    = & ~ -f(\cdot)_{j_0} \cdot \langle f(\cdot)_{j_0}, \A_{j_0, i} \rangle
\end{align*}
where the 1st step follows from basic calculus, the 2nd step follows from Part 4, the 3rd step follows from definition of $f(\cdot)_{j_0}$.

Using all of the results above, it holds that
\begin{align*}
   \frac{ \d f(\cdot)_{j_0} }{ \d x_i }  = & ~ f(\cdot)_{j_0} \circ \A_{j_0,i} - f(\cdot)_{j_0} \cdot \langle f(\cdot)_{j_0} , \A_{j_0,i}\rangle
\end{align*}

{\bf Proof of Part 6.}
It follows Part 5 directly.

{\bf Proof of Part 7.}
 For simplicity of writing proofs, we use $(\cdot)$ to denote $(x)$.

Following the definition of $c$ in Definition~\ref{def:c}, it holds that
\begin{align}\label{eq:c_x:}
    c(\cdot)_{j_0,i_0} := \langle f(\cdot)_{j_0}, v \rangle - E_{j_0,i_0}
\end{align}

Thus it holds that
\begin{align*}
        \frac{ \d c(\cdot)_{j_0,i_0} }{ \d x_i } 
        = & ~ \frac{ \d (\langle f(\cdot)_{j_0}, h(y)_{i_0} \rangle - E_{j_0,i_0}) }{ \d x_i } \\
        = & ~ \frac{ \d \langle f(\cdot)_{j_0}, h(y)_{i_0} \rangle }{ \d x_i } \\
        = & ~ \langle  f(\cdot)_{j_0} \circ \A_{j_0,i},  h(y)_{i_0} \rangle - \langle  f(\cdot)_{j_0} ,  h(y)_{i_0} \rangle \cdot \langle f(\cdot)_{j_0}, \A_{j_0,i} \rangle,
    \end{align*}
    where the 1st step is because of Eq.~\eqref{eq:c_x:}, the 2nd step is from $\frac{\d E_{j_0,i_0}}{\d x_i} = 0$, and the 3rd step is followed by {\bf Part 4}.

{\bf Proof of Part 8.}
 For simplicity of writing proofs, we use $(\cdot)$ to denote $(x)$.
Following the definition of $L(\cdot)$ in Definition~\ref{def:l}, it holds that
\begin{align}\label{eq:l_x:}
    L(\cdot)_{j_0,i_0} = 0.5 c(\cdot)_{j_0,i_0}^2 
\end{align}

Thus, we have
\begin{align*}
    \frac{\d L(\cdot)_{j_0,i_0}}{\d x_i} 
    = & ~ \frac{\d (0.5 c(\cdot)_{j_0,i_0}^2)}{\d x_i} \\
    = & ~ c(\cdot)_{j_0,i_0} \frac{\d c(\cdot)}{\d x_i} \\
    = & ~ c(\cdot)_{j_0,i_0} \cdot (\langle  f(\cdot)_{j_0} \circ \A_{j_0,i},  h(y)_{i_0} \rangle - \langle  f(\cdot)_{j_0} ,  h(y)_{i_0} \rangle \cdot \langle f(\cdot)_{j_0}, \A_{j_0,i} \rangle),
\end{align*}
where the 1st step is followed by the Eq.~\eqref{eq:l_x:}, the 2nd step is due to the chain rule, the last step followed by {\bf Part 5}.

\end{proof}

\section{Time for Computation}\label{sec:app_time}

In Section~\ref{sec:compute_f_h}, we show the calculation of $f$ (Similarly as Section~\ref{sec:app_gradient}, we still ignore the $1/d$ factor here) and $h$. In Section~\ref{sec:compute_c}, we show the way we calculate $c$ in straightforward way. In Section~\ref{sec:compute_q} and Section~\ref{sec:compute_p}, we define two artificial  functions $p$ and $q$, and show how to compute them. In Section~\ref{sec:compute_gradient}, we provide the way to re-write the gradient in an elegant way. In Section~\ref{sec:compute_together}, we finally put these all together and find the running time of our algorithm. 

\subsection{Compute \texorpdfstring{$f$}{} and \texorpdfstring{$h$}{}}\label{sec:compute_f_h}

\begin{lemma}[Computing $f$ and $h$]\label{lem:compute_f_h}
Suppose the following objects are given
\begin{itemize}
    \item Let $f(x)$ be defined as Definition~\ref{def:f}
    \item Let $h(y)$ be defined as Definition~\ref{def:h}
\end{itemize}
Then, we have 
\begin{itemize}
    \item $f(x)$ can be calculated in time of $\Tmat(n,d,n) + \Tmat(n,d,d)$
    \item $h(y)$ can be calculated in time of $\Tmat(n,d,d)$
\end{itemize}
\end{lemma}
\begin{proof}
Note that 
\begin{align*}
f(x)  = D^{-1} \exp(A_1 X A_2^\top )
\end{align*}
and
\begin{align*}
 D = \diag( \exp(A_1 X A_2^\top ) {\bf 1}_n )
\end{align*}
We firstly compute $\exp(A_1 X A_2^\top )$, this takes time of $\Tmat(n,d,d)$ and $\Tmat(n,d,n)$. 

Then we can compute $D$, which takes $O(n^2)$ time.

Then we can compute $D^{-1} \exp(A_1 X A_2^\top )$, this takes $O(n^2)$ time.

Thus, the overall time is
\begin{align*}
    & ~ \Tmat(n,d,d) + \Tmat(n,d,n) + O(n^2) \\
    = & ~ O( \Tmat(n,d,d) + \Tmat(n,d,n) )
\end{align*}

Note that $h(y) = A_3 Y$ which takes time of $\Tmat(n,d,d)$. 

Thus, the proof is completed.
\end{proof}

\subsection{Compute \texorpdfstring{$c$}{}}\label{sec:compute_c}

\begin{lemma}[Computing $c$]\label{lem:compute_c}
Suppose the following objects are given
\begin{itemize}
    \item $B \in \R^{n \times d}$
    \item $f(x) \in \R^{n \times n}$ is given
    \item $h(y) \in \R^{n \times d}$ is given,
\end{itemize}
Then one can compute $c(x) \in \R^{n \times d}$ in $\Tmat(n,n,d)$ time.
\end{lemma}
\begin{proof}
Based on Definition of $c(x) \in \R^{n \times d}$ which is
\begin{align*}
     c(x) = f(x) h(y) - E
\end{align*}
Computing $f(x) h(y)$ takes time of $\Tmat(n,n,d)$, and calculating $f(x) h(y) - E$ takes time of $O(nd)$.

Thus, finally, overall time is 
\begin{align*}
\Tmat(n,n,d) + O(nd).
\end{align*}

\end{proof}

\subsection{Computation for \texorpdfstring{$q$}{}}\label{sec:compute_q}

We will define $q$, and then explain how to calculate $q$.
\begin{definition}\label{def:q}
Define $c(x) \in \R^{n \times d}$ as in Definition~\ref{def:c}. Define $h(y) \in \R^{n \times d}$ as in Definition~\ref{def:h}.

We define $q(x) \in \R^{n \times n}$ as
\begin{align*}
    q(x) : = \underbrace{ c(x) }_{n \times d} \underbrace{ h(y)^\top }_{d \times n}
\end{align*}
Then we use $q(x)_{j_0}^\top$ to denote the $j_0$-th row of $q(x) \in \R^{n \times n}$.
\end{definition}

\begin{lemma}\label{lem:compute_q}
If it holds that
\begin{itemize}
    \item Suppose $c(x)\in \R^{n \times d}$ is given
    \item Suppose $h(y) \in \R^{n \times d}$ is given
\end{itemize}
Then, we can compute $q(x)$ in the time of $O(\Tmat(n,n,d))$.
\end{lemma}
\begin{proof}
Recall that $q(x) = c(x) h(y)^\top$. Thus it takes time of $\Tmat(n,d,n) = O(\Tmat(n,n,d))$.
\end{proof}

\subsection{Computation for \texorpdfstring{$p(x)$}{}}\label{sec:compute_p}

Let us firstly define $p$, and then we can show how to construct it.
\begin{definition}\label{def:p}
For every index $j_0 \in [n]$, we define $p(x)_{j_0} \in \R^n$ as
\begin{align*}
p(x)_{j_0} := ( \diag( f(x)_{j_0} ) - f(x)_{j_0} f(x)_{j_0}^\top) q(x)_{j_0}.
\end{align*}
We define $p(x) \in \R^{n \times n}$ in the sense that $p(x)_{j_0}^\top$ is the $j_0$-th row of $p(x)$.
\end{definition}

\begin{lemma}\label{lem:compute_p}
If the below requirements are holding that
\begin{itemize}
    \item Suppose $f(x)\in \R^{n \times n}$ is given
    \item Suppose $q(x) \in \R^{n \times n}$ is given
\end{itemize}
Then, we can compute $q(x)$ in $O(n^2)$ time.
\end{lemma}
\begin{proof}
Since $\diag(f(x)_{j_0})$ is a diagonal matrix and $f(x)_{j_0} f(x)_{j_0}^\top$ is a rank-one matrix, we know that $p(x)_{j_0} \in \R^n$ can be computed in $O(n)$, for each $j_0 \in [n]$. Thus we can construct matrix $p(x) \in \R^{n \times n}$ in $n \times O(n) = O(n^2)$ time in total.
\end{proof}

\subsection{Analyze the closed form of gradient}\label{sec:compute_gradient}

\begin{lemma}[ ]\label{lem:compute_gradient}
Define the functions $f(x) \in \R^{n \times n}$, $c(x) \in \R^{n \times d}$,  $h(y) \in \R^{n \times d}$, $q(x) \in \R^{n \times n}$ and $p(x) \in \R^{n \times n}$ as in Definitions~\ref{def:f}, \ref{def:c}, \ref{def:h}, \ref{def:q} and \ref{def:p} respectively. $A_1, A_2 \in \R^{n \times d}$ are two given matrices. We define$\A = A_1 \otimes A_2$. Let $L(x)$ be defined as Definition~\ref{def:attention_optimization_loss}. Let $L(x)_{j_0,i_0}$ be defined as Definition~\ref{def:l}.
Then, we can show that $\frac{\d L(x)}{\d x} = \vect(A_1^\top p(x) A_2)$.
\end{lemma}
\begin{proof}

From the Lemma statement, we have
\begin{align}\label{eq:lxy_j0_i0}
    \frac{\d L(x,y)_{j_0,i_0}}{\d x_i} 
    =  c(x,y)_{j_0,i_0} \cdot (\langle  f(x)_{j_0} \circ \A_{j_0,i}, h(y)_{i_0} \rangle - \langle  f(x)_{j_0} , h(y)_{i_0} \rangle \cdot \langle f(x)_{j_0}, \A_{j_0,i} \rangle)
\end{align}

Note that by Fact~\ref{fac:circ_rules}, it holds that
\begin{align*}
    \langle  f(x)_{j_0} \circ \A_{j_0,i}, h(y)_{i_0} \rangle = \A_{j_0,i}^\top \diag(f(x)_{j_0}) h(y)_{i_0}
\end{align*}
and 
\begin{align*}
    \langle  f(x)_{j_0} , v \rangle \cdot \langle f(x)_{j_0}, \A_{j_0,i} \rangle
    = \A_{j_0,i}^\top f(x)_{j_0} f(x)_{j_0}^\top h(y)_{i_0}
\end{align*}

Therefore, Eq.~\eqref{eq:lxy_j0_i0} becomes
\begin{align}\label{eq:rewrite_single_loss_gradient}
    \frac{\d L(x)_{j_0,i_0}}{\d x_i} 
    = & ~ c(x,y)_{j_0,i_0} \cdot (\A_{j_0,i}^\top \diag(f(x)_{j_0}) h(y)_{i_0} - \A_{j_0,i}^\top f(x)_{j_0} f(x)_{j_0}^\top h(y)_{i_0}) \notag \\
    = & ~ c(x,y)_{j_0,i_0} \cdot \A_{j_0,i}^\top ( \diag(f(x)_{j_0}) - f(x)_{j_0} f(x)_{j_0}^\top)h(y)_{i_0},
\end{align}
where the 2nd step follows from simple algebra.

Recall the way we define $q(x)_{j_0}$ (see Definition~\ref{def:q}). 
\begin{align}\label{eq:q_xy_j0}
q(x)_{j_0} := \sum_{i_0=1}^d c(x)_{j_0,i_0} h(y)_{i_0}.
\end{align}

Recall that $p(x)_{j_0} \in \R^n$ is define as Definition~\ref{def:p}, 
\begin{align}\label{eq:p_xy_j0}
p(x)_{j_0} := ( \diag( f(x)_{j_0} ) - f(x)_{j_0} f(x)_{j_0}^\top) q(x)_{j_0}.
\end{align}

It holds that
\begin{align*}
    & ~ \frac{\d L(x)}{\d x} \\
    = & ~ \sum_{j_0=1}^n \sum_{i_0=1}^d \frac{\d L(x)_{j_0,i_0} }{ \d x } \\
    = & ~ \sum_{j_0=1}^n \sum_{i_0=1}^d \underbrace{ c(x)_{j_0,i_0} }_{ \mathrm{scalar} } \cdot \underbrace{ \A_{j_0}^\top }_{d^2 \times n} \underbrace{ ( \diag( f(x)_{j_0} ) - f(x)_{j_0} f(x)_{j_0}^\top ) }_{n \times n} \underbrace{ h(y)_{i_0} }_{n \times 1}  \\
    = & ~ \sum_{j_0=1}^n \A_{j_0}^\top ( \diag( f(x)_{j_0} ) - f(x)_{j_0} f(x)_{j_0}^\top) q(x)_{j_0} \\
    = & ~ \sum_{j_0=1}^n \A_{j_0}^\top p(x)_{j_0} \\
    = & ~ \vect( A_1^\top p(x) A_2 )
\end{align*}
where the 1st step is because of Definition~\ref{def:attention_optimization_loss}, the 2nd step is based on Eq.~\eqref{eq:rewrite_single_loss_gradient}, the 3rd step is followed by Eq.~\eqref{eq:q_xy_j0}, the 4th step is due to Eq.~\eqref{eq:p_xy_j0}, and the last step uses tensor-trick.

 \end{proof}

\subsection{Putting it together}\label{sec:compute_together}

\begin{lemma}[Attention gradient computation, formal version of Lemma~\ref{lem:gradient:informal}]\label{lem:gradient:formal}

If it holds that
\begin{itemize}
    \item Define $A_1, A_2, A_3, E \in \R^{n \times d}$. Define $X, Y \in \R^{d \times d}$ to be several input fixed matrices.
    \item Let $X, Y \in \R^{d \times d}$ denote matrix variables (we will compute gradient with respect to $X$ )
    \begin{itemize}
        \item For easy of writing, we also use vector variables $x \in \R^{d^2 \times 1}$ and $y \in \R^{d^2 \times 1}$, i.e., $\vect(X) = x$.
    \end{itemize}
    \item Let $g = \frac{\d L(X)}{\d x} \in \R^{d^2}$ (where $L(X)$ is defined as Definition~\ref{def:attention_optimization_loss})
\end{itemize}
Then we can show that gradient $g \in \R^{d^2}$ can be computed in $\Tmat(n,d,n) + \Tmat(n,d,d)$ time.
\end{lemma}
\begin{proof}

Step 1. we compute $f(x)$, $h(y)$. This takes $O(\Tmat(n,n,d) + \Tmat(n,d,d)) $ time due to Lemma~\ref{lem:compute_f_h}.

Step 2. we compute $c(x)$. This takes time of $O(\Tmat(n,n,d) + \Tmat(n,d,d)) $ due to Lemma~\ref{lem:compute_c}.

Step 3. we compute $q(x)$. This take time of $O(\Tmat(n,n,d))$ due to Lemma~\ref{lem:compute_q}.

Step 4. we compute $p(x)$. This take time of $O(n^2)$ due to Lemma~\ref{lem:compute_p}.

Step 5. using Lemma~\ref{lem:compute_gradient}, we know that gradient is equivalent to $\vect(A_1^\top p(x) A_2)$. Suppose $A_1^\top \in \R^{d \times n}, p(x) \in \R^{n \times n}, A_2 \in \R^{n \times d}$ are given, then it can be calculated in time of $O(\Tmat(n,n,d) + \Tmat(n,d,d) )$.

Thus, overall running for computing gradient is
\begin{align*}
O( \Tmat(n,d,d) + \Tmat(n,d,n))
\end{align*}
time.
\end{proof}

\section{Fast Running Time via Polynomial Method}\label{sec:app_fast_time}

Recall that in the previous section, for convenience of computing the derivative, we ignoreed the $d$ factor in $f$. That factor $d$ doesn't impact the running time of our algorithms since it is just a rescaling factor. To apply the tools from previous work \cite{as23}, we will now reconsider the $1/d$ factor in $f$. In Section~\ref{sec:low_rank_f}, we will show how to efficiently and explicitly construct a low rank representation for $f$. In Section~\ref{sec:low_rank_c}, we show how to create a low rank construction for $c(x)$. In Section~\ref{sec:low_rank_q}, Section~\ref{sec:low_rank_p1} and Section~\ref{sec:low_rank_p2}, we further give low rank presentations for $q(x), p_1(x), p_2(x)$. In Section~\ref{sec:low_rank_final}, we prove our final algorithmic result by putting everything together.

\subsection{Low rank representation to \texorpdfstring{$f$}{} }\label{sec:low_rank_f}
Using \cite{as23}'s polynomial method result, we are able to obtain the following low-rank representation result,
\begin{lemma}[Section 3 of \cite{as23}]\label{lem:low_rank_f}
For any $B = o(\sqrt{\log n})$, there exists a $k_1 = n^{o(1)}$ such that: Let $A_1, A_2 \in \R^{n \times d}$ be two matrices and $X \in \R^{d \times d}$ be a square matrix. It holds that $\| A_1^\top X \|_{\infty} \leq B, \| A_2 \|_{\infty} \leq B$, then there are two matrices $U_1, V_1 \in \R^{n \times k_1}$ such that $\| U_1 V_1^\top - f(x) \|_{\infty} \leq \epsilon/\poly(n)$. Here $f(x) = D^{-1} \exp(A_1XA_2^\top /d)$ and we define $D = \diag( \exp(A_1 X A_2^\top /d) {\bf 1}_n )$.  Moreover, these matrices $U_1, V_1$ can be explicitly constructed in $n^{1+o(1)}$ time.
\end{lemma}

\subsection{Low rank representation to \texorpdfstring{$c$}{}}\label{sec:low_rank_c}

\begin{lemma}\label{lem:low_rank_c}
Let $d = O(\log n)$. Assume that each number in the $n \times d$ matrices $E$ and $h(y)$ can be written using $O(\log n)$ bits. Let $n \times d$ matrix $c(x)$ be defined as Definition~\ref{def:c}. Then, there are two matrices $U_1, V_1 \in \R^{n \times k_1}$ we have $\| U_1 V_1^\top h(y) - E - c(x) \|_{\infty} \leq \epsilon / \poly(n)$.
\end{lemma}
\begin{proof}

We can show that
\begin{align*}
    \| U_1 V_1^\top h(y) - E - c(x) \|_{\infty}
    = & ~ \| U_1 V_1^\top h(y) - E - f(x) h(y) + E \|_{\infty} \\
    = & ~ \| ( U_1 V_1^\top -f(x) ) h(y) \|_{\infty} \\
    \leq & ~ \epsilon /\poly(n)
\end{align*}
where the first step follows from $c(x) = f(x) h(y) - E$.

\end{proof}

\subsection{Low rank representation to \texorpdfstring{$q$}{}}\label{sec:low_rank_q}
\begin{lemma}\label{lem:low_rank_q}
Let $k_2 = n^{o(1)}$.
Define $c(x) \in \R^{n \times d}$ to be as in Definition~\ref{def:c}. Define $h(y) \in \R^{n \times d}$ to be as in Definition~\ref{def:h}. 
Assume that $q(x) := h(y) c(x)^\top \in \R^{n \times n}$. There are two matrices $U_2, V_2 \in \R^{n \times k_2}$ such that $\| U_2 V_2^\top - q(x) \|_{\infty} \leq \epsilon / \poly(n)$. The matrices $U_2, V_2$ can be explicitly constructed in $n^{1+o(1)}$ time.
\end{lemma}
\begin{proof}

We define $\wt{q}(x)$ to be the approximation of $q(x)$.

From Lemma~\ref{lem:low_rank_c}, we know that $U_1 V_1^\top h(y) - E$ is a good approximation to $c(x)$.

Then we should pick in this way $\wt{q}(x) = h(y) ( U_1 V_1^\top h(y) - E  )^\top $.

Now, let us turn $\wt{q}(x)$ into some low-rank representation 
\begin{align*}
    \wt{q}(x) = \underbrace{ h(y) }_{n \times d} \underbrace{ h(y)^\top }_{d \times n} \underbrace{ V_1 }_{n \times k_1} \underbrace{ U_1^\top }_{k_1 \times n} - \underbrace{ h(y) }_{n \times d} \underbrace{ E^\top }_{d \times  n}
\end{align*}
It is obvious that we should can first compute $h(y)^\top V_1$ which only takes $n^{1+o(1)}$ time. Then since all the low rank matrices are known, then we can explicitly construct $U_2 , V_2 \in \R^{n \times k_2}$ where $k_2 = \max\{d,k\} + d  = n^{o(1)}$.

For controlling the error, we can show
\begin{align*}
   \|  \wt{q}(x) - q(x) \|_{\infty} 
   = & ~ \| h(y) (U_1V_1^\top h(y) ) - E)^\top - h(y) c(x)^\top \|_{\infty} \\
   \leq & ~ d \cdot \| h(y) \|_{\infty} \cdot \| U_1V_1^\top h(y) ) - E - c(x) \|_{\infty} \\
   \leq & ~ \epsilon / \poly(n)
\end{align*}
Thus, we complete the proof.
\end{proof}

\subsection{Low rank representation to \texorpdfstring{$p_1(x)$}{}}\label{sec:low_rank_p1}

\begin{lemma}\label{lem:low_rank_p1}
Let $k_1 = n^{o(1)}$. Let $k_2 = n^{o(1)}$.
Assume that $p_1(x) := f(x) \circ q(x)$. Assume $U_1, V_1 \in \R^{n \times k_1}$ approximates the $f(x)$ such that $\| U_1 V_1^\top -f (x) \|_{\infty} \leq \epsilon/\poly(n)$. Assume $U_2, V_2 \in \R^{n \times k_2}$ approximates the $q(x) \in \R^{n \times n}$ such that $\| U_2 V_2^\top -q(x) \|_{\infty} \leq \epsilon/\poly(n)$. Then there are matrices $U_3, V_3 \in \R^{n \times k_3}$ such that $\| U_3 V_3^\top - p_1(x) \|_{\infty} \leq \epsilon / \poly(n)$. The matrices $U_3, V_3$ can be explicitly constructed in $n^{1+o(1)}$ time.
\end{lemma}
\begin{proof}

We choose $U_3 = U_1 \oslash U_2$ and $V_3 = V_1 \oslash V_2$. This can be computed in $n^{1+o(1)}$ time.

For easy of writing proofs, we call $\wt{f}(x) = U_1 V_1^\top$ and $\wt{q}(x) = U_2 V_2^\top$.

Using Fact~\ref{fac:olash_folklore}, we know that
\begin{align*}
\| U_3 V_3^\top - p_1(x) \|_{\infty}
\leq & ~ \| U_3 V_3^\top - f(x) \circ q(x) \|_{\infty} \\
= & ~ \| (U_1 \oslash U_2)  (V_1 \oslash V_2)^\top - f(x) \circ q(x) \|_{\infty} \\
= & ~ \| (U_1 V_1^\top) \circ  ( U_2 V_2^\top ) -  f(x) \circ q(x) \|_{\infty} \\
= & ~ \| \wt{f}(x) \circ \wt{q}(x) - f(x) \circ q(x) \|_{\infty} \\
= & ~ \| \wt{f}(x) \circ \wt{q}(x) - \wt{f}(x) \circ q(x)  + \wt{f}(x) \circ q(x) - f(x) \circ q(x) \|_{\infty} \\
\leq & ~ \| \wt{f}(x) \circ \wt{q}(x) - \wt{f}(x) \circ q(x)  \|_{\infty} + \| \wt{f}(x) \circ q(x) - f(x) \circ q(x) \|_{\infty} \\
\leq & ~ \epsilon/ \poly(n)
\end{align*}
where the 1st step follows from the way we define $p_1(x)$, the 2nd step follows from the way we define $U_3$ and $V_3$, the 3rd step follows from Fact~\ref{fac:olash_folklore}, the 4th step follows from the way we define $\wt{f}(x)$ and $\wt{q}(x)$, the 5th step follows from simple algebra, the 6th step follows by triangle inequality, and the last step follows by that entries are bounded and $\| \wt{f}(x) - f(x) \|_{\infty} \leq \epsilon / \poly(n)$ (Lemma assumption) 
and  $\| \wt{q}(x) - q(x) \|_{\infty} \leq \epsilon / \poly(n)$ (Lemma assumption) 

\end{proof}

\subsection{Low rank representation \texorpdfstring{$p_2(x)$}{}}\label{sec:low_rank_p2}

\begin{lemma}\label{lem:low_rank_p2}
Let $k_1 = n^{o(1)}$. Let $k_2 = n^{o(1)}$. Let $k_4 = n^{o(1)}$.
Assume that $p_2(x)$ is an $n \times n$  where $j_0$-th column $p_2(x)_{j_0} = f(x)_{j_0} f(x)_{j_0}^\top q(x)_{j_0}$ for each $j_0 \in [n]$. Assume $U_1, V_1 \in \R^{n \times k_1}$ approximates the $f(x)$ such that $\| U_1 V_1^\top -f (x) \|_{\infty} \leq \epsilon/\poly(n)$. Assume $U_2, V_2 \in \R^{n \times k_2}$ approximates the $q(x) \in \R^{n \times n}$ such that $\| U_2 V_2^\top -q(x) \|_{\infty} \leq \epsilon/\poly(n)$. Then there are matrices $U_4, V_4 \in \R^{n \times k_4}$ such that $\| U_4 V_4^\top - p_2(x) \|_{\infty} \leq \epsilon / \poly(n)$. The matrices $U_4, V_4$ can be explicitly constructed in $n^{1+o(1)}$ time.
\end{lemma}

\begin{proof}

We define a local vector function $r(x) \in \R^{n}$ where $r(x)_{j_0}$ is $f(x)_{j_0} q(x)_{j_0}$. Let $\wt{r}(x)$ denote the approximation of $r(x)$.

Note that $(U_1 V_1)_{j_0,*}^\top$ is a good approximation to $f(x)_{j_0}$.

Note that $(U_2 V_2)_{j_0,*}^\top$ is a good approximation to $q(x)_{j_0}$.

Let $\wt{r}(x)_{j_0 }: = \langle \wt{f}(x)_{j_0} , \wt{q}(x)_{j_0} \rangle = (U_1 V_1)_{j_0,*} \cdot (U_2 V_2)_{j_0,*}^\top$.

For the computation side, we firstly compute $V_1 V_2^\top$. This takes $n^{1+o(1)}$ time.

Next, we we have
\begin{align*}
\wt{r}(x)_{j_0} 
= & ~ (U_1 V_1)_{j_0,*} \cdot (U_2 V_2)_{j_0,*}^\top \\
= & ~ \underbrace{ (U_1)_{j_0,*} }_{1 \times k_1} \underbrace{ V_1 V_2^\top }_{k_1 \times k_2}   \underbrace{ (U_2)_{j_0,*}^\top }_{k_2 \times 1}
\end{align*}
Once the $V_1 V_2^\top$ are pre-computed, the above step only takes $O(k_1k_2)$ time. Since there $n$ coordinates, so the overall time is still $O(n k_1 k_2) = n^{1+o(1)}$.

Let $\wt{f}(x) = U_1 V_1^\top$ denote the approximation of $f(x)$. Then we just use $\wt{f}(x)$ and $\wt{r}(x)$ to approximate $p_2(x)$ in the following sense, let $\wt{p}_2(x) = \wt{f}(x) \diag(\wt{r}(x))$. Since $\wt{f}(x)$ has low rank representation, and $\diag(\wt{r}(x))$ is a diagonal matrix, then it is obvious how to construct $U_4$ and $V_4$. Basically $U_4 = U_1$ and $V_4  = \diag(\wt{r}(x) ) V_1$.

Now, we need to control the error, we have
\begin{align*}
    \| U_4 V_4^\top - p_2(x) \|_{\infty}
    = & ~ \| \wt{p}_2(x) - p_2(x) \|_{\infty} \\
    = & ~ \max_{j_0 \in [n] } \| \wt{f}(x)_{j_0} \wt{r}(x)_{j_0} - f(x)_{j_0} r(x)_{j_0} \|_{\infty} \\
    = & ~ \max_{j_0 \in [n] } \| \wt{f}(x)_{j_0} \wt{r}(x)_{j_0} - \wt{f}(x)_{j_0} r(x)_{j_0} + \wt{f}(x)_{j_0} r(x)_{j_0}   - f(x)_{j_0} r(x)_{j_0} \|_{\infty} \\
    \leq & ~ \max_{j_0 \in [n] } \| \wt{f}(x)_{j_0} \wt{r}(x)_{j_0} - \wt{f}(x)_{j_0} r(x)_{j_0} \|_{\infty} + \| \wt{f}(x)_{j_0} r(x)_{j_0}   - f(x)_{j_0} r(x)_{j_0} \|_{\infty} 
\end{align*}
where the 2nd step follows follows from definition of $p_2(x)$ and $\wt{p}_2(x)$.

For the first term, we have
\begin{align*}
    \max_{j_0 \in [n] } \| \wt{f}(x)_{j_0} \wt{r}(x)_{j_0} - \wt{f}(x)_{j_0} r(x)_{j_0} \|_{\infty}
    \leq & ~ \max_{j_0 \in [n] }  \| \wt{f}(x)_{j_0} \|_{\infty} \cdot | \wt{r}(x)_{j_0} - r(x)_{j_0}| \\
    \leq & ~ \epsilon / \poly(n)
\end{align*}
For the second term, we have
\begin{align*}
    \max_{j_0 \in [n] } \| \wt{f}(x)_{j_0} r(x)_{j_0}   - f(x)_{j_0} r(x)_{j_0} \|_{\infty} \leq & ~ \max_{j_0 \in [n]} \| \wt{f}(x)_{j_0} - f(x)_{j_0} \|_{\infty} \cdot | r(x)_{j_0}| \\
    \leq & ~ \epsilon/ \poly(n)
\end{align*}

Using the three equations we obtained above, the proof is completed.
\end{proof}

\subsection{Fast Computation in Almost Linear Time}\label{sec:low_rank_final}

\begin{theorem}[Main result, formal version of Theorem~\ref{thm:mainalg}]\label{thm:mainalg:formal}
Assuming the entries of $A_1, A_2, X, A_3, Y, E$ are represented using $O(\log n)$ bits, there is a $n^{1+o(1)}$ time algorithm to solve $\mathsf{AAttLGC}(n, d = O(\log n), B = o(\sqrt{\log n} ))$ (see Definition~\ref{def:AAttLGC}) up to $1/\poly(n)$ accuracy. In particular, our algorithm outputs a gradient vector $\wt{g} \in \R^{d^2}$ such that $\| \frac{\d L}{\d x} - \wt{g} \|_{\infty} \leq 1/ \poly(n)$.
\end{theorem}
\begin{proof}
Recall definition of $n \times n$ matrices $p(x)$ (Definition~\ref{def:p}), $p_1(x)$ (see Lemma~\ref{lem:low_rank_p2}) and $p_2(x)$ (Lemma~\ref{lem:low_rank_p1}), it is straightforward that
\begin{align*}
    p(x) = p_1(x) - p_2(x).
\end{align*}

Using Lemma~\ref{lem:low_rank_f}, Lemma~\ref{lem:low_rank_c}, Lemma~\ref{lem:low_rank_q}, we know that assumptions in Lemma~\ref{lem:low_rank_p1} and Lemma~\ref{lem:low_rank_p2} are holding, so that we can use Lemma~\ref{lem:low_rank_p1} and Lemma~\ref{lem:low_rank_p2} to obtain that
\begin{itemize}
    \item $p_1(x)$ has approximate low rank representation $U_3, V_3$, let $\wt{p}_1(x)$ denote $U_3 V_3^\top$
    \item $p_2(x)$ has approximate low rank representation $U_4, V_4$, let $\wt{p}_2(x)$ denote $U_4 V_4^\top$
\end{itemize}
All of the Lemmas~\ref{lem:low_rank_f}, \ref{lem:low_rank_c}, \ref{lem:low_rank_q}, \ref{lem:low_rank_p1} and \ref{lem:low_rank_p2} are taking $n^{1+o(1)}$ time.

According to the proof for the Lemma~\ref{lem:compute_gradient}, we have that
\begin{align*}
    \frac{L(X)}{\d x} = \vect( A_1^\top p(x) A_2 )
\end{align*}

Thus, we firstly compute $A_1^\top U_3 V_3^\top A_2$,
\begin{itemize}
    \item We compute $A_1^\top U_3 \in \R^{d \times k_3}$, this takes $n^{1+o(1)}$ time
    \item We compute $V_3^\top A_2 \in \R^{k_3 \times d}$, this takes $n^{1+o(1)}$ time
    \item Compute $(A_1^\top U_3 ) \cdot (V_3^\top A_2) $, this takes $d^2 n^{o(1)}$ time
\end{itemize}
Second, we can compute $A_1^\top U_4 V_4^\top A_2$,
\begin{itemize}
    \item We compute $A_1^\top U_4 \in \R^{d \times k_4}$, this takes $n^{1+o(1)}$ time
    \item We compute $V_4^\top A_2 \in \R^{k_4 \times d}$, this takes $n^{1+o(1)}$ time
    \item Compute $(A_1^\top U_4 ) \cdot (V_4^\top A_2) $, this takes $d^2 n^{o(1)}$ time
\end{itemize}
So, overall running time is still $n^{1+o(1)}$.

We have
\begin{align*}
\| \frac{\d L(X)}{\d x} - \wt{g} \|_{\infty}
= & ~ \| \vect(A_1^\top p(x) A_2) -  \vect(A_1^\top \wt{p}(x) A_2) \|_{\infty}\\
= & ~ \| A_1^\top p(x) A_2 -  A_1^\top \wt{p}(x) A_2 \|_{\infty}\\
= & ~ \| A_1^\top (p_1(x) - p_2(x)) A_2 -  A_1^\top ( \wt{p}_1(x) - \wt{p}_2(x) ) A_2 \|_{\infty}\\
\leq & ~ \| A_1^\top ( p_1(x) - \wt{p}_1(x) ) A_2 \|_{\infty} + \| A_1^\top ( p_2(x) - \wt{p}_2(x) ) A_2 \|_{\infty} \\
\leq & ~ \| A_1 \|_{\infty} \| A_2 \|_{\infty} \cdot n^2 \cdot (  \| p_1(x) - \wt{p}_1(x) \|_{\infty} + \| p_2(x) - \wt{p}_2(x) \|_{\infty} ) \\
\leq & ~ \epsilon / \poly(n)
\end{align*}
where the 4th step follows from triangle inequality, the last step follows from entries in $A_1,A_2$ are bounded, and $\| p_1(x) - \wt{p}_1(x) \|_{\infty} \leq \epsilon/\poly(n)$, $\| p_2(x) - \wt{p}_2(x) \|_{\infty} \leq \epsilon/\poly(n)$ .

Picking $\epsilon = 1/\poly(n)$, we have the proof completed.
\end{proof}

\section*{Acknowledgments}
The authors would like to thank Yichuan Deng for helpful discussions.

\bibliographystyle{alpha}
\bibliography{ref}





\end{document}